\documentclass{article}

\usepackage{arxiv}

\usepackage{microtype}
\usepackage{graphicx}
\usepackage{subcaption}
\usepackage{booktabs} 
\usepackage[hidelinks]{hyperref}

\usepackage{amsmath}
\usepackage{amssymb}
\usepackage{mathtools}
\usepackage{amsthm}
\usepackage[capitalize,noabbrev]{cleveref}

\theoremstyle{plain}

\theoremstyle{definition}

\theoremstyle{remark}

\usepackage[textsize=tiny]{todonotes}
\usepackage{times}
\usepackage{latexsym}
\graphicspath{ {figures/} }
\usepackage{verbatim}
\usepackage{multirow}
\usepackage{array}
\usepackage{pifont}
\usepackage{xcolor, colortbl}
\usepackage{soul}
\usepackage{algorithm}
\usepackage{float}
\usepackage[T1]{fontenc}
\usepackage[utf8]{inputenc}
\definecolor{LightLightGray}{gray}{0.9}
\usepackage[round]{natbib}

\title{How Graph Structure and Label Dependencies\\Contribute to Node Classification\\in a Large Network of Documents}

\author{
 Pirmin Lemberger \\
  onepoint\\
  29 rue des Sablons, 75016, Paris (France)\\
  \texttt{p.lemberger@groupeonepoint.com} \\
   \And
 Antoine Saillenfest\\
  onepoint\\
  29 rue des Sablons, 75016, Paris (France) \\
  \texttt{a.saillenfest@groupeonepoint.com} \\
}

\begin{document}
\maketitle
\begin{abstract}
We introduce a new dataset named \mbox{\textit{WikiVitals}} which contains a large graph of 48k mutually referred Wikipedia articles classified into 32 categories and connected by 2.3M edges. Our aim is to rigorously evaluate the contributions of three distinct sources of information to the label prediction in a semi-supervised node classification setting, namely the content of the articles, their connections with each other and the correlations among their labels. We perform this evaluation using a Graph Markov Neural Network which provides a theoretically principled model for this task and we conduct a detailed evaluation of the contributions of each sources of information using a clear separation of model selection and model assessment. One interesting observation is that including the effect of label dependencies is more relevant for sparse train sets than it is for dense train sets.
\end{abstract}

\section{Introduction}
Graph neural networks (GNN) have become a tool of choice when modeling datasets whose observations are not i.i.d. but are comprised of entities interconnected according to a graph of relations \citep{yang2016revisiting, defferrard2016convolutional, kipf2017semi}. They can be used either for graph classification, like molecule classification \citep{dobson2003distinguishing, borgwardt2005protein} or for node classification, like document classification in a citation network \citep{sen2008collective}. The most common task is the semi-supervised node classification in which unlabeled nodes of a given subset are to be classified using a distinct subset of labeled nodes from the same graph \citep{defferrard2016convolutional, kipf2017semi}. This is the task on which we focus in this paper. 

A number of architectures have been proposed over the years which deal with specific issues occurring with GNNs. Some combat over-smoothing (which is the tendency for deep GNNs to predict the same labels for all nodes) \citep{klicpera2018predict}, some deal with assortativity or heterophily (which refers to situations in which neighboring nodes are likely to have different labels) \citep{zhu2020h2GCN,zhu2021graph,bo2021beyond} and others still try to learn the connection weights from data using an appropriate attention mechanism \citep{velickovic2017graph}.  

Despite their diversity, these models all have one important limitation. Namely they assume that labels can be predicted independently for each node in the graph. In other words they neglect label dependencies altogether. More recently Graph Markov Neural Networks (GMNN) \citep{qu2019gmnn} were introduced as consistent probabilistic models which include label correlations in graphs in a natural way by combining the strength of GNNs and those of conditional random fields (CRF). 

Our aim in this work is to study how texts in a relatively large network of documents can be classified in a semi-supervised setting by leveraging three distinct sources of information. Beyond the content of a document, which is obviously the most relevant piece of information, we intend to rigorously evaluate the weaker but still relevant contributions of the network structure and the label dependencies. We use GMNNs for this purpose because they provide text representations which integrate those three sources. 

In a nutshell a GMNN uses two coupled GNNs. One is a mean field approximation which predict labels from text features as if the labels were uncorrelated. Another GNN, coupled to the first, models the label dependencies. This makes it easy to disentangle the three contributions. A graph agnostic baseline which only takes into account the content of the documents is obtained by collapsing a GMNN to a simple MLP which makes independent predictions for each document. Leveraging the graph structure but omitting label correlations is obtained by using only the mean field GNN which predicts uncorrelated labels. At last, a full-fledged GMNN will also include the label correlations.

The contribution of label correlations to the prediction accuracy of a GMNN, which is a small effect, was evaluated in \cite{qu2019gmnn} on the classical benchmark datasets Cora, Pubmed and Citeseer \citep{sen2008collective}. As Table \ref{tab:datasets_statistics} shows, these datasets however are rather small in terms of the number of links when compared to a graph which includes, say, a significant fraction of Wikipedia. Moreover the evaluation was done using one single public split per dataset as defined in \citet{yang2016revisiting}. Under these settings an improvement was demonstrated when comparing the GMNN model to existing baselines that do not account for label dependencies. 

However, as a number of recent works \citep{shchur2018pitfalls,errica2020fair} have pointed out, a fair evaluation of the performance of GNNs requires a procedure which performs a systematic randomization over train-validation-test set partitions and makes clear separation between model selection and model assessment.

Our aim in this paper is to subject GMNN to such a rigorous performance analysis on a new, relatively large graph of documents with topical labels named \textit{WikiVitals} that we created for that purpose. For completeness we also perform the same thorough analysis on the classical benchmark datasets Cora, Citeseer and Pubmed.

In summary, our contributions\footnote{Code and data are available at: \newline\url{https://github.com/ToineSayan/node-classification-and-label-dependencies/}} are:
\begin{itemize}

    \item We introduce a new dataset of interconnected documents named \mbox{\textit{WikiVitals}} extracted from the English Wikipedia. Compared to the classical benchmark datasets this is a relatively large graph comprising 48k nodes classified into 32 categories and connected by 2.3M edges.
    
    \item We apply the fair comparison procedure proposed in \citet{errica2020fair} to a GMNN. So far only graph classification models had been evaluated in this manner.
    
    \item We evaluate the respective contributions to the classification accuracy of the content of the \mbox{\textit{WikiVitals}} articles, the graph structure and the correlations between their labels. We distinguish between sparse train sets and dense train sets, where labeled nodes make up a respectively a small and a large fraction of the nodes in the network. In both cases our result confirm that taking into account the label correlation provides a statistically significant improvement but also that the effect is much stronger for sparse train sets than for dense train sets.
    
\end{itemize}

\section{Related Work}
\subsection{Evaluating Performance of GNNs}
As the authors of \citet{shchur2018pitfalls} point out, the evaluations of GNN models are almost never conducted in a rigorous manner. On the one hand, many experiments cannot be reproduced due to the lack of a precise definition of the evaluation process. On the other hand, they argue that using a single split, usually the one defined in the paper that introduces a new benchmark dataset, is insufficient to guarantee the existence of a significant difference between the accuracy of two competing GNN architectures. The authors thus suggest standardizing the choice of hyperparameters and randomizing over many train-validation-test splits. Then they search for a set of hyperparameters that optimizes the average performance over those splits. Surprisingly, they find that the simplest architectures like GCNs \citep{defferrard2016convolutional, kipf2017semi} often perform better for the semi-supervised node classification task than the more sophisticated models \citep{monti2017geometric, velickovic2017graph}. 

In our work we follow a still more rigorous accuracy assessment that was originally proposed in \citet{errica2020fair} as a SOTA evaluation procedure for the graph classification task. For a given model we search for the best hyperparameters on a per split basis and then average the accuracy estimations of those optimized models over splits. This allows for a fair assessment when comparing two models in the sense that it guarantees that a practitioner who randomly chooses a split, trains her model on the train set, optimizes its hyperparameters on the validation set and estimates the accuracy on the test set will obtain an estimation that is reliable for comparing models such as a MLP, a GCN or a GMNN.

\subsection{Modelling Label Dependency in GNNs}
Prior to the recent advent of GNNs a number of works have attempted to include label dependencies using various heuristics. Label propagation is such an early attempt where a cost function balances the penalty for predicting the wrong labels with the requirement that node labels should vary smoothly \citep{zhou2003learning,zhu2005semi}. 

Dataset specific methods have also been proposed. As far as classifying Wikipedia articles is concerned, authors in \cite{viard2020classifying} use a simple GNN whose weights are empirically adjusted depending on the similarity of the labels of neighboring nodes. Although these approaches had some empirical success \citep{huang2020combining}, the lack of a sound probabilistic foundation makes it difficult to analyze why they fail or succeed. In particular they do not clearly distinguish the contributions of the node features, the graph structure and the label correlations to the prediction accuracy. For this reason we decided to avoid using topological node features like node degree, betweenness or assortativity \citep{newman2003mixing,newman2005measure,blondel2008fast} in our work to make this distinction clearer.

GNNs are a good fit for computing distributed node representations that merge the information supplied by the node features, like the content of a document for instance, with the local structure of the graph in its vicinity. Each such representation is then used for predicting the label for that node independently of those of the other nodes. CRFs on the other hand come in handy for describing label correlations. However performing exact inference is hard due to the difficulty of computing their partition function. GMNN propose an elegant solution to this conundrum by using two ordinary GNNs which are coupled when trained with the Expectation Maximization (EM) algorithm. Their performance was evaluated in \citet{qu2019gmnn} in the usual way using public partitions of small citation networks \citep{sen2008collective} and without accounting for the robustness of this evaluation when using different splits which is essential for a fair evaluation \citep{shchur2018pitfalls,errica2020fair}. Our goal is to provide a detailed and rigorous evaluation for a large real world graph of documents.

\subsection{Classifying Wikipedia Articles}
Wikipedia articles provide rich textual content from which  representations can be computed, for example using informative $n$-grams. The hyperlinks define a graph whose nodes are the articles. Several datasets including Squirrel and Chameleon \citep{Benedek2021}\footnote{\url{http://snap.stanford.edu/data/wikipedia-article-networks.html}} have been created from Wikipedia and are commonly used to evaluate various GNN architectures. This is also the case for the \mbox{\textit{WikiVitals}} dataset we introduce in this article.

The semantic of the labels is different in each dataset. The labels of Squirrel and Chameleon for instance are based on monthly traffic data (acquired through the meta-data of the articles) which was discretized into 3 or 5 categories \citep{bo2021beyond}. In \citet{viard2020classifying}, the labels were defined outside of Wikipedia. None of these datasets however exploit a thematic classification resulting from a consensus among Wikipedia editors as does the list of vital articles of Wikipedia\footnote{\url{https://en.wikipedia.org/wiki/Wikipedia:Vital\_articles/Level/5}} that we use to define our \mbox{\textit{WikiVitals}} dataset (section \ref{sec:wikivitals}). Furthermore, the classification is imbalanced and contains categories with very few representatives which makes it a difficult classification problem. 

A common feature of Wikipedia datasets (Squirrel, Chameleon as well as \mbox{\textit{WikiVitals}}) is that they are more disassortative \citep{newman2003mixing} than classical graph datasets\footnote{The notion of heterophily is also commonly used and means that most neighbors $n'$ of a node $n$ with label $\mathbf{y}_n$ have a label $\mathbf{y}_{n'}$ which differs from $\mathbf{y}_{n}$.}. This makes them particularly interesting as benchmarks for a node classification task because basic models like GCNs show struggle in such disassortative contexts \citep{bo2021beyond}. Some recent models like H$_{2}$GCN or FAGCN have been proposed to overcome this problem and show better performance in those contexts \citep{zhu2020h2GCN,bo2021beyond}. 
We experiment with both GCN and FAGCN (section \ref{subsec:DS_and_Settings}).

\section{Adapting the Fair Comparison Method to GMNN}
\subsection{Training GMNNs}
\label{AdaptToGMNN}
In order to explain how we perform a rigorous evaluation of GMNNs in the next subsection we briefly recall how these models are defined and how they are trained, referring to \citet{qu2019gmnn} for a thorough presentation.

We consider a graph $G=(V,E,\mathbf{x}_V)$ where $V$ denotes the set of nodes, $E$ the set of edges and $\mathbf{x}_V:=\{\mathbf{x}_n\}_{n\in V}$ the set of features associated to each node $n$. We assume that we are given the one-hot encoded labels (for $K$ categories) $\mathbf{y}_L:=\{\mathbf{y}_n\}_{n\in L}$ for the nodes in a subset $L\subset V$ and the features $\mathbf{x}_V$ of all nodes. The task we consider is the prediction of the labels $\mathbf{y}_U$ of the remaining unlabeled nodes in $U=V\setminus L$. The GMNN model does two things. First, it specifies a model for the joint probability $p_\phi(\mathbf{y}_L, \mathbf{y}_U|\mathbf{x}_V)$ which accounts for correlations between neighboring nodes. Second, it describes a practical training procedure, based on the EM algorithm, for finding the parameters $\phi$ which maximize a variational lower bound on the marginal likelihood $p_\phi(\mathbf{y}_L|\mathbf{x}_V)$ over the observed labels.

Training a GMNN requires defining two ordinary GNNs. The first one, denoted by $\text{GNN}_\phi$, where $\phi$ is the set of its parameters, describes the distribution $p_\phi(\mathbf{y}_n|\mathbf{y}_{\mathrm{NB}(n)},\mathbf{x}_V)$ over individual node labels $\mathbf{y}_n$ given the labels of the neighboring nodes denoted by $\mathrm{NB}(n)$ and the node features $\mathbf{x}_V$. It is specified in the usual manner by a softmax applied on a $d$-dimensional node embedding $\mathbf{h}_{\phi,n}$, read off from the last layer of $\text{GNN}_\phi$, multiplied by a $K\times d$ learnable matrix $W_\phi$
\begin{equation}
	p_\phi(\mathbf{y}_n|\mathbf{y}_{\mathrm{NB}(n)},\mathbf{x}_V)=\mathrm{Cat}(\mathbf{y}_n|\mathrm{softmax}(W_\phi\mathbf{h}_{\phi,n})).
\end{equation}
A second GNN, that we denote by $\text{GNN}_\theta$, defines a mean-field variational distribution meant to approximate the posterior $p_\phi(\mathbf{y}_U|\mathbf{y}_L,\mathrm{x}_V)$ in the EM algorithm. It is defined nodewise in a similar way
\begin{equation}
	q_\theta(\mathbf{y}_n|\mathbf{x}_V)=
	\mathrm{Cat}(\mathbf{y}_n|\mathrm{softmax}(W_\theta\mathbf{h}_{\theta,n})).
\end{equation}
Intuitively $\text{GNN}_\theta$ is a model that completely neglects correlations among labels. These predictions are then adjusted by $\text{GNN}_\phi$ which accounts for the correlations between the labels of neighboring nodes, these in turn will correct $\text{GNN}_\theta$ within an EM cyclic training procedure. The training process uses the following two objective functions. One is for updating $\theta$ while holding $\phi$ fixed:
\begin{equation}
\label{Otheta}
	O_{\theta} = \sum_{n\in U} \mathbb{E}_{p_\phi(\mathbf{y}_n|\hat{\mathbf{y}}_{\mathrm{NB}(n)},\mathbf{x}_V)}
	[\log q_\theta(\mathbf{y}_n|\mathbf{x}_V)]	+ \sum_{n\in L} \log q_\theta (\mathbf{y}_n|\mathbf{x}_V), 
\end{equation}
where $\hat{\mathbf{y}}_{n}$ denotes the ground truth label $\mathbf{y}_{n}$ if $n\in L$ and is sampled from $q_\theta(\mathbf{y}_{n}|\mathbf{x}_V)$ if $n\in U$. Using the same notations, the other objective function used for optimizing $\phi$ while holding $\theta$ fixed is:
\begin{equation}
\label{Ophi}
	O_\phi = \sum_{n\in V} \log p_\phi(\hat{\mathbf{y}}_n|\hat{\mathbf{y}}_{\mathrm{NB}(n)},\mathbf{x}_V).
\end{equation}
The first step of training a GMNN is to initialize $q_\theta$ by maximizing the last term in (\ref{Otheta}) for $\theta$. This corresponds to an ordinary GNN trained without accounting for label correlations. The accuracy of this initial $q_\theta$ model will thus provide a baseline to compare with the full GMNN model. Second, fix $\theta$ and optimize $\phi$ in (\ref{Ophi}), this is the $M$-step. At last, optimize (\ref{Otheta}) for $\theta$ while holding $\phi$ fixed, this is the $E$-step. Repeat the $M$ and $E$ step until convergence. Experience shows that $q_\theta$  consistently yields a better predictor than $p_\phi$ \citep{qu2019gmnn}.

\subsection{Fair Comparison of GMNNs}
\label{sec:FC}
\begin{figure*}
    \centering
    \includegraphics[width=\textwidth]{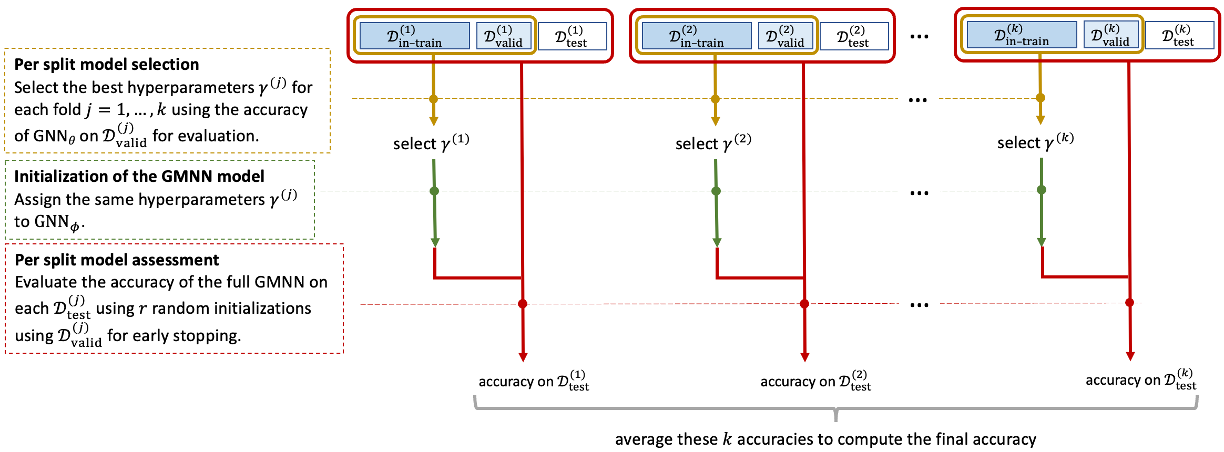}
    \caption{The fair evaluation procedure for GNNs and its adaptation for GMNN uses $k$ train/validation/test splits $\mathcal{D}^{(i)}_{\mathrm{in-train}}, \mathcal{D}^{(i)}_{\mathrm{valid}}, \mathcal{D}^{(i)}_{\mathrm{test}}$ which are created from $k$ stratified folds $\mathcal{F}_i$.}
    \label{fig:FC_GMNN}
\end{figure*}
Recall our main goals. First we want to ascertain whether accounting for label correlations using a GMNN model leads to a predictor with a higher accuracy in a statistically significant way when applied to a large real-world dataset like \mbox{\textit{WikiVitals}}. We thus describe in detail how we adapt the approach in \citet{errica2020fair} to GMNNs to perform this rigorous evaluation. For the same dataset we also wish to compare a structure agnostic baseline like an MLP with ordinary GNNs like GCN or FAGCN which leverage the graph structure but do not account for label correlations. For those simpler models we apply their method without modifications.

A crude evaluation would proceed by partitioning the dataset of labeled  articles $\mathcal{D}=((\mathbf{x}_1,\mathbf{y}_1),\dots,(\mathbf{x}_N,\mathbf{y}_N))$ into three disjoint sets: a train set $\mathcal{D}_{\mathrm{train}}$, a validation set $\mathcal{D}_{\mathrm{valid}}$ for selecting the optimal hyperparameters $\gamma^*$ and a test set $\mathcal{D}_{\mathrm{test}}$ to evaluate the accuracy of that optimal model. Unfortunately such a simple procedure was shown to be so unstable that changing the partition could totally scramble relative ranking of various GNN architectures \citep{shchur2018pitfalls,errica2020fair}.

The main requirement for reliable evaluations is to clearly separate model selection from model assessment. For that we split the dataset $\mathcal{D}$ the usual way, using $k$ disjoint stratified\footnote{In the context of this paper, stratified sampling of nodes means that the distribution of labels in the sampled set is the same as in the set of nodes from which it was sampled.} folds $\mathcal{F}_1,\dots,\mathcal{F}_k$. Then $k$ different train and test sets are defined as: 
\[
	\mathcal{D}^{(i)}_{\mathrm{train}}:=\bigcup_{j\neq i}
	\mathcal{F}_j, \:\:\:\:
	\mathcal{D}^{(i)}_{\mathrm{test}}:=\mathcal{F}_i,
	\:\:\:\: i=1,\dots,k.
\]
Each train set is in turn split into an inner train set and a validation set:
\[
	\mathcal{D}^{(i)}_{\mathrm{train}}:=
	\mathcal{D}^{(i)}_{\mathrm{in-train}}\cup
	\mathcal{D}^{(i)}_{\mathrm{valid}},
	\:\:\:\: i=1,\dots,k.
\]
Figure \ref{fig:FC_GMNN} shows the fair evaluation procedure adapted to GMNN.
\newline\newline
\textbf{Model selection} is the choice of an optimal set of hyperparameters. In contrast with the standard $k$-fold cross-validation procedure we perform this selection separately for each $\mathcal{D}^{(i)}_{\mathrm{train}}$. As in \cite{qu2019gmnn} we optimize the hyperparameters of the mean-field network $\mathrm{GNN}_\theta$ and then use these same optimal values as hyperparameters for the other $\mathrm{GNN}_\phi$ network which encodes label correlations. More precisely, the $\mathrm{GNN}_\theta$ model is trained on $\mathcal{D}^{(i)}_{\mathrm{in-train}}$ using $\mathcal{D}^{(i)}_{\mathrm{valid}}$ as a holdout set for selecting the hyperparameters $\gamma^{(i)}$ which minimize the loss.
\newline\newline
\textbf{Model assessment} evaluates the accuracy of the whole GMNN (which couples $\mathrm{GNN}_\theta$ and $\mathrm{GNN}_\phi$) separately on each test set $\mathcal{D}^{(i)}_{\mathrm{test}}$ using the optimal hyperparameters $\gamma^{(i)}$ for split $i$. The final accuracy assessment is then the average of these results over the $k$ splits.   More precisely, the full-fledged GMNN model is trained using the EM algorithm on each $\mathcal{D}^{(i)}_{\mathrm{in-train}}$ set using $\mathcal{D}^{(i)}_{\mathrm{valid}}$ for early stopping and is then evaluated on $\mathcal{D}^{(i)}_{\mathrm{test}}$.

Note that performing a fair evaluation of the model after completion of the initial training of GNN$_\theta$ and before entering the EM optimization corresponds to a fair evaluation of a plain GNN which thus requires no additional computation.

\section{Experiment}
\subsection{Creating \textit{WikiVitals}}
\label{sec:wikivitals}
\begin{table}[ht!]
    \footnotesize
    \centering
    \begin{tabular}{llr} 
         \specialrule{.2em}{.1em}{.1em} 
           & Class name & \#articles \\ 
         \cmidrule{1-3}
            & \hspace*{-0.04\textwidth}\textit{Arts} & \\
            & 01-Arts & 3310 \\
            & \hspace*{-0.04\textwidth}\textit{Biological and health sciences}& \\
            & 02-Animals & 2396 \\
            & 03-Biology &  886 \\
            & 04-Health & 791 \\
            & 05-Plants & 608 \\
            & \hspace*{-0.04\textwidth}\textit{Everyday life} & \\
            & 06-Everyday life & 1191 \\
            & 07-Sports, games and recreation & 1231 \\
            & \hspace*{-0.04\textwidth}\textit{Geography} & \\
            & 08-Cities  & 2030 \\
            & 09-Countries & 1386 \\
            & 10-Physical & 1902 \\
            & \hspace*{-0.04\textwidth}\textit{History} & \\
            & 11-History  & 2979 \\
            & \hspace*{-0.04\textwidth}\textit{Mathematics} & \\
            & 12-Mathematics & 1126 \\
            & \hspace*{-0.04\textwidth}\textit{People} & \\
            & 13-Artists, musicians, and composers & 2310 \\
            & 14-Entertainers, directors, producers, and screenwriters & 2342 \\
            & 15-Military personnel, revolutionaries, and activists & 1012 \\
            & 16-Miscellaneous & 1186\\
            & 17-Philosophers, historians, political and social scientists & 1335 \\
            & 18-Politicians and leaders & 2452 \\
            & 19-Religious figures & 500 \\
            & 20-Scientists, inventors, and mathematicians & 1108 \\
            & 21-Sports figures & 1210 \\
            & 22-Writers and journalists & 2120 \\
            & \hspace*{-0.04\textwidth}\textit{Philosophy and religion} & \\
            & 23-Philosophy and religion & 1408 \\
            & \hspace*{-0.04\textwidth}\textit{Physical sciences} & \\
            & 24-Astronomy & 886 \\
            & 25-Basics and measurement & 360 \\
            & 26-Chemistry & 1207 \\
            & 27-Earth science & 849\\
            & 28-Physics & 988 \\
            & \hspace*{-0.04\textwidth}\textit{Society and social sciences} & \\
            & 29-Culture & 2075 \\
            & 30-Politic and economic & 1825 \\
            & 31-Social studies & 355\\
            & \hspace*{-0.04\textwidth}\textit{Technology} & \\
            & 32-Technology & 3148 \\
         \specialrule{.2em}{.1em}{.1em} 
    \end{tabular}
    \caption{The 32 node labels used in the \mbox{\textit{WikiVitals}} dataset classified by topics.}
    \label{tab:wikivitals_labels}
\end{table}

We created \textit{WikiVitals} as a disassortative document-document network from 48512 vital Wikipedia articles extracted from a complete Wikipedia dump dated April 2022. Nodes correspond to vital Wikipedia articles. Node features are sparse binary bag-of-words representations of the articles. Each of the 4000 features in these representations corresponds to the presence or absence of an informative unigram or bigram in the introduction, title or section titles of the article. Edges correspond to the hyperlinks between articles within the corpus of vital articles. 

Vital articles have been selected by Wikipedia editors and categorized by topic. We extracted a 3-level hierarchy of topics and used the 32 intermediate topics within this hierarchy as labels assigned to each node in the graph. Each node was assigned a single label. Table \ref{tab:wikivitals_labels} shows a partial view of the topic hierarchy, focusing on the 32 intermediate categories we used in this paper\footnote{The top level of the hierarchy comprises 11 coarse topics, the middle level 32 topics and the finest level 251 topics.}. We relied on a dump of the English Wikipedia because it provides a frozen view of the English Wikipedia and because it contains the text of the articles.

We performed feature extraction on the abstracts, titles and headers for the vital articles. To pre-process this data, we ignored the stop words using  \texttt{nltk.corpus} and applied stemming using \texttt{Snowballstemmer} from \texttt{nltk}. Next, we extracted unigrams and bigrams with a frequency greater than $1 \times 10^{-3}$ from the abstracts and headers  and those with a frequency greater than $1 \times 10^{-4}$ from the titles. Finally, we retained the top 4000 features that were most predictive of the labels in the chi-squared sense.

\begin{table}
    \footnotesize
    \centering
    \begin{tabular}{lrrrrr}
         \specialrule{.2em}{.1em}{.1em} 
         Dataset & Assortativity & \#Nodes & \#Edges & \#Categories & \#Features \\ 
         \cmidrule{1-6}
         Cora & 0.771 & 2,708 & 5,429 & 7 & 1,433\\
         Citeseer & 0.675 & 3,327 & 4,732 & 6 & 3,703 \\
         Pubmed & 0.686 & 19,717 & 44,338 & 3 & 500\\
         WikiVitals & 0.204 & 48,512 &  2,297,782 & 32 & 4,000\\
         \specialrule{.2em}{.1em}{.1em} 
    \end{tabular}
    \caption{Statistics of document graphs}
    \label{tab:datasets_statistics}
\end{table}

\subsection{Datasets and Settings}
\label{subsec:DS_and_Settings}
\textbf{Datasets:} 
Beside the WikiVitals dataset we also performed a fair evaluation on the three well-known assortative citation network datasets: Cora, Citeseer, and Pubmed. Edges in these networks represent citations between two scientific articles, node features $\mathbf{x}_n$ are a bag-of-words vector of the articles and labels $\mathbf{y}_n$ correspond to the fields of the articles. For all datasets, we treat the graphs as undirected. Statistics of these datasets are shown in Table \ref{tab:datasets_statistics}.
\newline\newline
\textbf{Training Details:} All baseline models (MLP, GCN and FAGCN) were reimplemented using PyTorch with two layers (input representations $\rightarrow$ hidden layer $\rightarrow$ output layer). For all models, $L^2$-regularization is performed on all layers, dropout is applied on input data and on all layers. For GCN and FAGCN, we used the so-called \textit{renormalization trick} of the adjacency matrix \citep{kipf2017semi}. For FAGCN, the number of propagations \citep{bo2021beyond} is set to 2 in order to limit the aggregation of information to nodes located at a maximum distance of 2. For GMNN we use the annealing sampling method with factor set to 0.1 \citep{qu2019gmnn}, the number of EM-loops is set to 10. To train the $\mathrm{GNN}_\phi$ network we used both the labels $\hat{\mathbf{y}}_{n'}$ of neighboring nodes $n'\in\mathrm{NB}(n)$ and node features $\mathbf{x}_V$ as defined in (\ref{Ophi}) to predict $\mathbf{y}_n$.

We use the same training procedure for all models. For all datasets, node features are binarized and then normalized (unit $L^1$-norm) before training. We used the Adam optimizer \citep{kingma2015adam} with default parameters and no learning rate decay, the same maximum number of training epochs, an early stopping criterion and a patience hyperparameter (see appendix \ref{sec:app} for more detail). Validation loss used for early stopping is evaluated at the end of each epoch. All model parameters (convolutional kernel coefficients for FAGCN, weight matrices for all models) are initialized and optimized simultaneously (weights are initialized according to Glorot and biases initialized to zero). In all cases we use full-batch training (using all nodes in the train set every epoch).
\newline\newline
\textbf{Fair evaluation setup:} 
During the assessment phase, we perform $r=20$ training runs with different random initialization of the weights to smooth out possibly bad configurations.

For each dataset, we followed the best practices advocated in \citet{errica2020fair} to pre-calculate stratified splits $(\mathcal{D}^{(j)}_{\mathrm{in-train}}, \mathcal{D}^{(j)}_{\mathrm{valid}}, \mathcal{D}^{(j)}_{\mathrm{test}}),\: j=1,\dots,k$ of the entire set of nodes with respective ratios of 81\%, 9\% and 10\%. In the sequel the sets $\mathcal{D}^{(j)}_{\mathrm{in-train}}$ will be referred to as dense train sets.

To enable convenient comparison with existing work we have created two more sets of splits whose train sets are sparse. As a reminder, the evaluation of GNNs as well as GMNN for Cora, Citeseer and Pubmed was classically performed using the Planetoid splits \citep{yang2016revisiting} or similarly constructed sparse splits composed of 20 nodes per category randomly selected from the whole dataset \citep{shchur2018pitfalls, qu2019gmnn,bo2021beyond}. 
To construct splits with sparse train sets we independently extracted two subsets $\mathcal{D}^{(j)}_{\mathrm{sparse-balanced}}$ and $\mathcal{D}^{(j)}_{\mathrm{sparse-stratified}}$ from each $\mathcal{D}^{(j)}_{\mathrm{in-train}},\: j=1,\dots,k$. Each contains $20\times K$ nodes (where $K$ is the number of categories). 

Each $\mathcal{D}^{(j)}_{\mathrm{sparse-balanced}}$ is constructed by selecting 20 nodes of each category from $\mathcal{D}^{(j)}_{\mathrm{in-train}}$. In the sequel these sets will be referred to as sparse balanced train sets in the sense that each category is represented equally in each of them. 

Each $\mathcal{D}^{(j)}_{\mathrm{sparse-stratified}}$ is constructed by selecting nodes from $\mathcal{D}^{(j)}_{\mathrm{in-train}}$ in a stratified way. We shall denote these sets as sparse stratified train sets. 

Rigorous model selection phases imply performing extensive grid searches over the set of hyperparameter, which is computationally very expensive. We have therefore implemented our own evolutionary grid search algorithm which discovers suitable hyperparameters by using the validation accuracy to guide the evolution. Such an algorithm computes a suitable configuration by exploring a small portion of the hyperparameter set \citep{young2015optimizing}.

\section{Results} 
Quantitative results for the node classification task applied to our \mbox{\textit{WikiVitals}} dataset and to the classical Cora, Citeseer and Pubmed datasets are shown in Table \ref{tab:results_FC_sparse} for sparse train sets and in Table \ref{tab:results_FC_dense} for dense train sets. The reported accuracies are the averages over the $k$ splits before and after the EM phases of the GMNN model. Each of these average accuracies was computed with $r$ different initializations to obtain an estimate of their variance which is written in brackets in the tables. When we say that the improvement of an accuracy is significant we mean that its significance level is better than $p<0.05$ \footnote{We performed a $t$-test with the alternate hypothesis is that the accuracy after the EM phase is higher than before.}. Table \ref{tab:results_FC_sparse} shows that the results for stratified and balanced sampling in the sparse train sets are almost equal.

\subsection{Contribution of the Graph Structure}
The results in Tables \ref{tab:results_FC_sparse} and \ref{tab:results_FC_dense} confirm that taking into account the graph structure provides a statistically significant performance boost for the node classification task for all datasets, regardless of whether the train set is dense or sparse. 

One important point to notice though is that because \mbox{\textit{WikiVitals}} is a disassortative dataset it is the FAGCN model which performs best. Actually when using a dense train set for \mbox{\textit{WikiVitals}} a simple MLP performs better than the basic GCN, thus confirming the importance to select an architecture which is adapted to the level of assortativity of the graph.

For the classical datasets on the other hand, both GCN and FAGCN models outperform the use an MLP which only takes into account node features disregarding the graph structure. 
\subsection{Contribution of the Label Dependencies}
The first observation is that leveraging label dependencies for \mbox{\textit{WikiVitals}} is significant whether the train sets is sparse or dense even though the impact is much stronger for a sparse train set than for a dense one. We interpret this
as a consequence that the information supplied by the correlations among labels is more useful in sparse setting where it compensates for the small number of nodes available for training. On the other hand, when the train set is dense, the information for making accurate predictions is more likely to be supplied by the features of a large number of nodes.

Leveraging label dependencies also significantly benefits the Cora datset for both sparse and dense train sets. For the dense case Pubmed benefits a clear increase in accuracy only when using FAGCN while Citeseer benefits no statistically significant increase in accuracy. The cases for which there is no significant improvement from taking into account label correlations with a GMNN are denoted in italics in Table \ref{tab:results_FC_dense}.
\begin{table*}[]
    \footnotesize
    \centering
    \resizebox{\textwidth}{!}{
    \setlength\tabcolsep{2.5pt}
    \begin{tabular}{l>{\columncolor[gray]{0.9}}r>     
    {\columncolor[gray]{0.9}}r rrr rrr}
         \specialrule{.2em}{.1em}{.1em} 
          & \multicolumn{2}{>{\columncolor[gray]{0.9}}c}{Wikivitals} 
          & \multicolumn{2}{c}{Cora} 
          & \multicolumn{2}{c}{Citeseer} 
          & \multicolumn{2}{c}{Pubmed} \\ 
          \cmidrule(l{2pt}r{2pt}){2-3} \cmidrule(l{2pt}r{2pt}){4-5} \cmidrule(l{2pt}r{2pt}){6-7} \cmidrule(l{2pt}r{2pt}){8-9}
          & balanced & stratified & balanced & stratified & balanced & stratified & balanced & stratified \\ 
         \cmidrule{1-9}
         MLP                            & 68.60 (0.92) & 69.35 (1.10) & 58.54 (3.98) & 58.32 (2.17) & 59.84 (3.54) & 58.35 (2.48) & 71.23 (2.85) & 70.39 (1.70) \\[4pt]
         GNN (base)                     & 70.64 (0.85) & 72.68 (1.17) & 80.78 (2.58) & 81.31 (2.16) & 69.05 (3.66) & 70.94 (2.16) & 80.20 (1.88) & 80.50 (2.38) \\[4pt]
         + GMNN 
                                        & \textbf{74.80 (1.18)} & \textbf{74.73 (1.36)} & \textbf{81.67 (3.00)} & \textbf{81.91 (2.21)} & \textbf{69.61 (3.96)} & \textbf{71.62 (2.20)} & \textbf{81.67 (1.32)} & \textbf{81.70 (2.45)} \\[4pt]
         \specialrule{.2em}{.1em}{.1em} 
    \end{tabular}
    }
    \caption{Sparse train sets: test accuracy averaged over $k$ splits reported in \%. Best results are highlighted. The base GNN is FAGCN for \mbox{\textit{WikiVitals}} and GCN for Cora, Citeseer and Pubmed.}
    \label{tab:results_FC_sparse}
\end{table*}
\begin{table}
    \footnotesize
    \centering
    \begin{tabular}{l>{\columncolor[gray]{0.9}}r rr rrr}
         \specialrule{.2em}{.1em}{.1em} 
          & WikiVitals & Cora & Citeseer & Pubmed \\ 
        \cmidrule{1-5}
         MLP & 86.55 (0.42) & 78.49 (2.39) & 75.02 (2.15) & 88.68 (0.86) \\[4pt]
         
         GCN & 72.74 (0.61) & 88.84 (2.39) & 77.24 (1.73) & 89.20 (0.86) \\
         + GMNN & 74.19 (0.42) & \textbf{89.26 (1.91)} & \textit{77.43 (1.70)}  & \textit{89.18 (0.84)} \\[4pt]
         
         FAGCN & 87.84 (0.32) & 88.87 (1.99) & 78.27 (3.53) & 90.23 (0.90) \\
         + GMNN & \textbf{87.92 (0.31)} & 89.08 (1.76) & \textit{\textbf{78.32 (3.64)}}  & \textbf{90.34 (0.88)} \\
         \specialrule{.2em}{.1em}{.1em} 
    \end{tabular}
    \caption{Dense train sets: test accuracy averaged over $k$ splits reported in \%. Best results are highlighted. Italics indicates cases for which there is no significant improvement from taking into account label correlations.}
    \label{tab:results_FC_dense} 
\end{table}
\section{Conclusion and Perspectives}
This paper introduces a new large and disassortative document-document graph dataset named \mbox{\textit{WikiVitals}} and adapts a fair comparison method of GNNs to GMNN to evaluate the contribution of three distinct sources of information for a semi-supervised node classification task: the node features which encodes the content the documents, the underlying graph structure and the label correlations. 

Taking into account label correlations was our main focus because this is the most subtle effect. We indeed confirmed that for \mbox{\textit{WikiVitals}} these correlations indeed provide a statistically significant improvement to the classification accuracy whether the nodes available for training are sparse or dense over the entire graph. The increase in accuracy is however much larger for sparse train sets. For the classical benchmark datasets Cora, Citeseer and Pubmed the improvement is statistically significant for sparse train sets, confirming existing results, but this improvement is much less obvious for dense train sets.

An interesting avenue for future research involves studying whether hierarchical categorization and multi-label classification could improve if we properly define an take into account correlations for these richer label structures. This would for instance allow us to deal with ambiguous situations in which some documents are assigned several labels, which often occurs in practice.

\section*{Broader Impact Statement}

The \textit{WikiVitals} we created could be useful in any research on large graphs of documents. We do not anticipate any negative societal consequences.

\section*{Acknowledgments and Disclosure of Funding}

This work was performed using HPC resources from GENCI–IDRIS (Grant 2021-AD011013266).

\vskip 0.2in
\bibliographystyle{plainnat} 
\bibliography{bibliography}

\appendix
\section{Hyperparameters}
\label{sec:app}

\textbf{Hyperparameters and search space}:
Grid search during model selection was performed over the following search space $\Gamma$:
\begin{itemize}
    \setlength{\itemsep}{0pt}
    \setlength{\parskip}{0pt}
    \setlength{\parsep}{0pt} 
  \item hidden dimension: [$8, 16, 32, 64$]
  \item input dropout: [$0.2, 0.4, 0.6, 0.8$]
  \item dropout: [$0.2, 0.4, 0.6, 0.8$]
  \item learning rate:  [1e-1, 5e-2, 1e-2, 5e-3, 1e-3, 5e-4, 1e-4]
  \item $L^2$-regularization strength:  [1e-1, 5e-2, 1e-2, 5e-3, 1e-3, 5e-4, 1e-4, 5e-5, 1e-5]
  \item $\epsilon$ (only for FAGCN):  [$0.1, 0.2, 0.3, 0.4, 0.5, 0.6, 0.7, 0.8, 0.9$]
\end{itemize}
For \mbox{\textit{WikiVitals}}, we use a reduced search space, hidden dimension was set to 64 and $L^2$-regularization strength to 1e-5, learning rate was in [1e-1, 5e-2], and $\epsilon$ in [$0.7, 0.8, 0.9$].
\newline\newline
\textbf{Training procedures for GNN models}: 
For all GNN model training:
\begin{itemize}
    \setlength{\itemsep}{0pt}
    \setlength{\parskip}{0pt}
    \setlength{\parsep}{0pt} 
  \item We train for a maximum of 1000 epochs.
  \item We use early stopping, patience is set to 200.
  \item There is no learning rate decay.
  \item An $L^2$-regularization is applied on all layers.
  \item All model parameters (convolutional kernel coefficients for FAGCN, weight matrices for all models) are optimized simultaneously.
  \item Once training has stopped, we reset the state of model parameters to the step with the lowest validation loss. 
\end{itemize}
For MLP and GCN, we use early stopping which stops the optimization if the validation loss does not decrease for 200 epochs. For FAGCN, we to stop the optimization if the validation loss and the validation accuracy do not decrease for 200 epoches. For GMNN training, we train models for 100 epoches and perform 10 EM-loops.
\newline\newline
\noindent\textbf{Evolutionary grid search}:
Our evolutionary algorithm maintains a randomly initialized population of 100 configurations of hyperparameters over generations. At each generation we retain between 2 and 50 configurations whose validation accuracy exceeds the population average for the next generation. New configurations are generated via a 2-pivot crossover, two configurations that have a better evaluation being more likely to be selected. A mutation step assigns a new value to a configuration hyperparameter with a probability 0.05 to promote exploration. Only configurations never seen before are added to complete the population at each generation. The number of generations is set to 10, a value beyond which the evaluation of the best configurations do not significantly increase anymore.

\end{document}